\documentclass{article}

\PassOptionsToPackage{numbers, compress}{natbib}

\usepackage[final]{neurips_2025}
\usepackage{graphicx}

\usepackage{algorithmic}
\usepackage{algorithm}
\usepackage{multirow}
\usepackage{stfloats}
\usepackage{tcolorbox}
\usepackage{subcaption}
\usepackage{amsthm}
\usepackage{authblk}
\usepackage[dvipsnames]{xcolor}
\usepackage{booktabs}

\newtheorem{theorem}{Theorem}
\newtheorem{lemma}{Lemma}

\usepackage{tikz}
\usepackage{amsmath,amssymb}

\usepackage[utf8]{inputenc} 
\usepackage[T1]{fontenc}    
\usepackage{hyperref}       
\usepackage{url}            
\usepackage{booktabs}       
\usepackage{amsfonts}       
\usepackage{nicefrac}       
\usepackage{microtype}      
\usepackage{xcolor}         

\title{Preprint: Toward Robust and Efficient ML-Based GPU Caching for Modern Inference}

\author{
Peng Chen$^{1}$,
Jiaji Zhang$^{1}$,
Hailiang Zhao$^{1\textcolor{green!80!black}{*}}$,
Yirong Zhang$^{1}$,
Shenyao Chen$^{1}$,
Jiahong Yu$^{1}$,
Xueyan Tang$^{2}$,
Yixuan Wang$^{3}$,
Hao Li$^{4}$,
Jianping Zou$^{4}$,
Gang Xiong$^{4}$,
Kingsum Chow$^{1}$,
Shuibing He$^{1}$,
Shuiguang Deng$^{1}$\thanks{Corresponding authors: Hailiang Zhao (\texttt{hliangzhao@zju.edu.cn}) and Shuiguang Deng (\texttt{dengsg@zju.edu.cn}).} \\

$^{1}$Zhejiang University \quad
$^{2}$Nanyang Technological University \quad \\
$^{3}$Nanjing University of Aeronautics and Astronautics \quad
$^{4}$Kuaishou
}

\begin{document}

\maketitle

\begin{abstract}

In modern GPU inference, cache efficiency remains a major bottleneck, and heuristic policies such as \textsc{LRU} can perform far worse than the offline optimum. Existing learning-based caching systems improve hit rates mainly through predictor design, but often follow learned predictions blindly, making performance unreliable when predictions are inaccurate. In contrast, emerging learning-augmented caching algorithms~\cite{pmlr-v80-lykouris18a,mitzenmacher2022algorithms} provide performance guarantees by carefully integrating predictions into caching policies, achieving both \emph{consistency} (near-optimality under perfect predictions) and \emph{robustness} (bounded worst-case performance under prediction errors). However, deployment remains challenging. A practical algorithm should satisfy strict time and space efficiency constraints, which some theoretical work overlooks, while also incurring low deployment overhead.

We propose learning-augmented LRU, a deployment-oriented learning-augmented caching algorithm that guarantees \emph{1-consistency} and \emph{$O(k)$-robustness}, incurs low time and space overhead, and maintains strong compatibility. We further build a GPU cache, called \textsc{LCR}, on top of learning-augmented LRU to benefit from its theoretical guarantees and translate them into practical performance. In experiments, \textsc{LCR} reduces P99 time-to-first-token (TTFT) by up to 28.3\% on LLM workloads and increases throughput by up to 24.2\% on deep learning recommendation (DLRM) workloads. Even with poor predictions, performance degrades gracefully and remains close to \textsc{LRU}, demonstrating robustness with practical value.

\end{abstract}

\section{Introduction}

In modern GPU inference systems, performance bottlenecks increasingly stem not from computation but from memory access~\cite{lee2021merci, lee2024infinigen, qiu2025flips, recasens2025mindmemorygapunveiling, yang2025gpu, cheng2025ecco}. Deep learning models continue to grow in scale and complexity, ranging from large language models (LLMs) that rely on extensive key-value (KV) cache reuse~\cite{zheng2024sglang} to deep learning recommendation models (DLRMs) with terabyte-scale embedding tables~\cite{jiang2021microrec}.

While high-bandwidth memory (HBM) on GPUs enables fast data access, its capacity is severely constrained compared to the working sets demanded by modern inference. Inefficient GPU caching translates directly into higher latency, reduced throughput, and degraded service-level agreement (SLA) compliance under real workloads. Meanwhile, inference access patterns are highly diverse. In LLM serving, prompt length, the number of conversation turns, and inter-request intervals vary widely, whereas DLRMs exhibit irregular and sparse embedding lookups. We defer a detailed discussion of GPU caching in production systems to Appendix~\ref{appendix:gpu_caching_in_production_systems}.

\begin{figure}[h]
    \centering
    \includegraphics[width=0.85\linewidth]{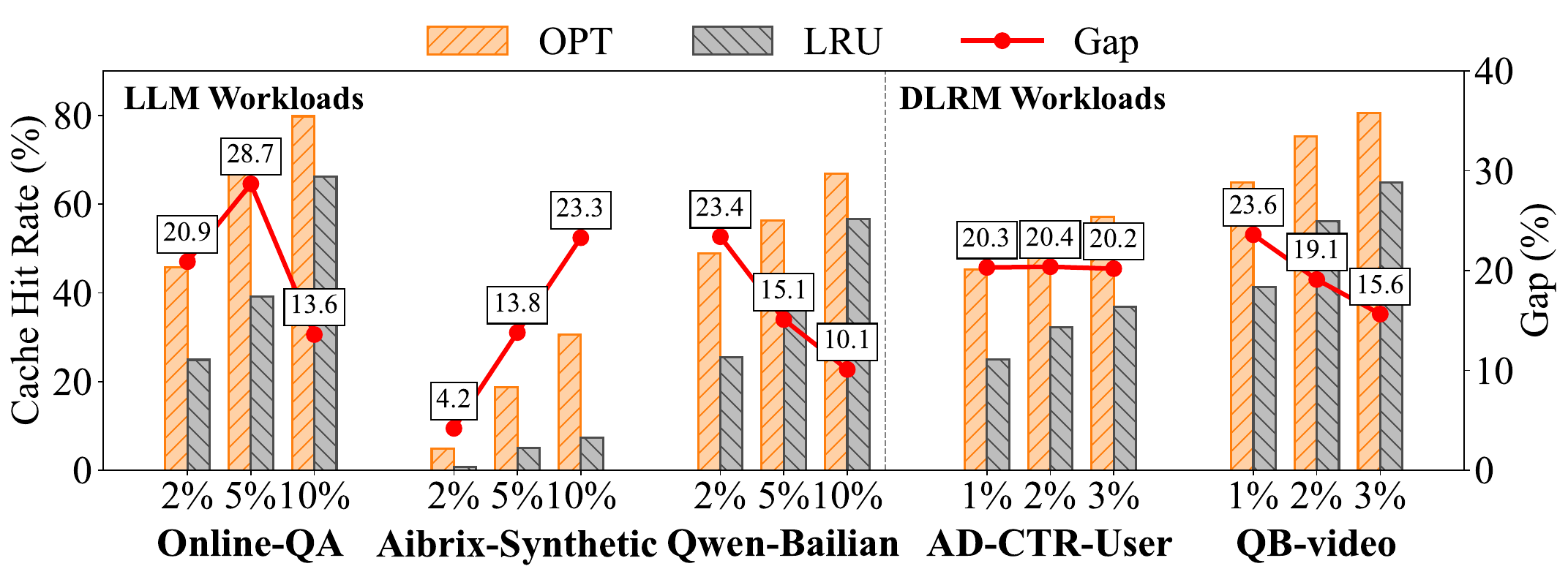}
    \caption{Cache hit rate of \textsc{LRU} and the offline optimum (\textsc{OPT}) under varying cache ratios.}
    \label{fig:cache_hit_rate_gap}
\end{figure}

Frameworks for both LLMs (e.g., SGLang~\cite{zheng2024sglang} and vLLM~\cite{kwon2023efficient}) and DLRMs (e.g., HugeCTR~\cite{wei2022gpu}) use \textsc{LRU} as the default eviction policy. While simple, \textsc{LRU} assumes recency predicts reuse, which often breaks under dynamic or scan-heavy access patterns~\cite{rodriguez2021learning}. Figure~\ref{fig:cache_hit_rate_gap} shows a sizable gap to the offline optimal (\textsc{OPT}): $15.6\%$--$23.6\%$ for DLRMs and up to $28.7\%$ for LLMs. Dataset details are deferred to the evaluation section.



The inefficiency of static heuristics can become a critical bottleneck, especially when cache misses trigger expensive recomputation or memory transfers, motivating ML-based caching approaches.

\paragraph{Existing learning-based caching systems and their limitations.}
To improve cache hit rates, recent work has explored learning-based caching~\cite{jain2016back, liu2020imitation, rodriguez2021learning, song2020learning, shi2019applying, zhou20253l, song2023halp, vietri2018driving, yan2021learning, yang2023gl, yang2023learned}, where systems such as \textsc{Glider}~\cite{shi2019applying}, \textsc{LRB}~\cite{song2020learning}, and \textsc{Parrot}~\cite{liu2020imitation} use machine learning (ML) to predict future accesses and guide eviction. However, these systems mainly focus on predictor design and deployment while largely overlooking cache policy design. Most simply adopt \textsc{FPB} (\textit{Follow Prediction Blindly}), which evicts based on predicted priorities and can fail severely when predictions are wrong. As a result, learning-based caches often degrade under mispredictions caused by data drift, cold starts, or adversarial inputs. A few works \cite{song2023halp, yang2023learned} take the conservative extreme, \textsc{HF} (\textit{Heuristic-Filtered}), which filters candidates down to a small set prioritized for eviction by a heuristic (e.g., \textsc{HALP} uses \textsc{LRU} to filter the candidate set down to four items) and then uses an ML model to rank them. This improves stability, but it can limit performance even when prediction accuracy remains high.

In GPU inference, a few poor evictions can trigger costly host-device transfers or recomputation, reducing DLRM throughput~\cite{wei2022gpu} and increasing LLM TTFT, especially in the tail~\cite{zheng2024sglang}. Thus, sharp performance drops under prediction errors remain a key barrier to production deployment~\cite{rodriguez2021learning, zhang2024sieve}. Further discussion of prior learning-based caching systems is deferred to Appendix \ref{appendix:existing_learning_based_caching_systems}.

\paragraph{Existing learning-augmented caching algorithms and the deployment gap.}
Motivated by the risk of blindly following predictions, a line of research on learning augmented algorithms has emerged.\footnote{See \url{https://algorithms-with-predictions.github.io/} for an overview.} These algorithms aim to perform strongly when predictions are accurate, which is captured by \emph{consistency}, while degrading gracefully and remaining robust when predictions are erroneous, which is captured by \emph{robustness}. Despite the theoretical guarantees offered by existing learning-augmented caching algorithms, deploying them for model inference remains challenging for several reasons: (1) algorithmic time complexity can be prohibitively high (e.g., \textsc{F\&R}~\cite{sadek2024algorithms}); (2) some rely on the randomized \textsc{Marker} heuristic~\cite{fiat1991competitive}, whose marking-based eviction is misaligned with the radix-tree prefix caching used in LLM serving, which typically restricts evictions to suffix items to preserve shared prefixes, thus requiring non-trivial adaptation (e.g., \textsc{PredictiveMarker}~\cite{pmlr-v80-lykouris18a}); and (3) others require maintaining multiple cache index structures to simulate the costs of different algorithms (e.g., \textsc{BlindOracle\&LRU}~\cite{wei2020better}), incurring substantial deployment overhead when the indexing is nontrivial, such as the \texttt{RadixTree} used in SGLang~\cite{zheng2024sglang}. We defer a detailed discussion of existing learning augmented caching algorithms to Appendix \ref{appendix:existing_learning-augmented_caching_algorithms_and_limitations}.

\subsection{Our contributions}

Motivated by the limitations of prior learning-based caching systems and inspired by learning-augmented caching principles, we seek to preserve the robustness of the classical \textsc{LRU} policy, which is widely used in inference frameworks and validated in industrial production, while exploiting accurate predictions to improve performance. To this end, we design a practical algorithm that adaptively leverages predictions based on their accuracy and validate it through real-world deployment.

\begin{itemize}
    \item We present learning-augmented LRU (\textsc{LARU}), a deployment-oriented learning augmented caching algorithm that augments \textsc{LRU} with predictions while dynamically adjusting its confidence in those predictions and in \textsc{LRU} priorities. Like \textsc{LRU}, \textsc{LARU} is compatible with most existing systems. When predictions are accurate, \textsc{LARU} fully exploits them to achieve optimal performance (\emph{1-consistency}). When prediction errors are large, it degrades gracefully and remains close to \textsc{LRU} (\emph{$O(k)$-robustness}, matching \textsc{LRU}'s competitive ratio up to constant factors), while maintaining $O(k)$ space overhead, matching \textsc{LRU}, and $O(n \log k)$ algorithmic time complexity, matching the lower bound for most algorithms that rank eviction candidates by priority. Here, $n$ represents the total number of requests and $k$ is the maximum number of items the cache can hold, independent of item size.
    \item Built on \textsc{LARU}, we develop \textsc{LCR}, a GPU cache that can be integrated into different model inference frameworks. We use it to validate \textsc{LARU} on real-world inference workloads. To the best of our knowledge, \textsc{LCR} is the first caching system to bring learning-augmented caching guarantees, including \emph{consistency} and \emph{robustness}, into practice, delivering strong performance with both accurate and inaccurate predictions.
    \item On LLM workloads, it reduces P99 TTFT by up to $10.7\%$ on Aibrix-Synthetic with Qwen2.5-32B and by $13.5\%$–$28.3\%$ on Online-QA with DeepSeek-R1-671B. On DLRM workloads, it improves SparseLengthsSum (SLS) throughput by up to $24.2\%$ on QB-Video, while maintaining robust performance even when predictions are inaccurate.
\end{itemize}


\section{Preliminaries}

We consider caching with fixed-size items, a common setting in GPU caching for model inference, including KV-cache vectors for LLMs and embedding vectors for DLRMs. The cache can store at most $k$ items. Requests arrive online as a sequence, with an item requested each time. A request to item $x$ is a cache hit if $x$ is already in cache; otherwise, it incurs a miss, and the requested item $x$ must be loaded, evicting a page if the cache is full, with the goal of minimizing the number of misses. The performance of an online caching algorithm $\text{ALG}$ is measured by its \emph{competitive ratio}, defined as
\[
CR(\text{ALG}) := \sup_{\mathcal{I}} \frac{\text{ALG}(\mathcal{I})}{\text{OPT}(\mathcal{I})},
\]
where $\mathcal{I}$ is a request sequence, and $\text{ALG}(\mathcal{I})$ and $\text{OPT}(\mathcal{I})$ denote the costs incurred by $\text{ALG}$ and the offline optimal algorithm $\text{OPT}$, respectively. We say that $\text{ALG}$ is $CR(\text{ALG})$-competitive. The cost is defined as the number of cache misses. \textsc{LRU} has a competitive ratio of $k$ \cite{sleator1985amortized}, which is optimal for deterministic algorithms.

For learning-augmented caching algorithms, the \emph{consistency} of $\text{ALG}$ is its competitive ratio under perfectly accurate predictions, while its \emph{robustness} is its competitive ratio under completely inaccurate predictions. Thus, the best possible consistency is $1$, meaning it matches the offline optimum under perfect predictions, and the robustness of any deterministic learning-augmented algorithm cannot be better than $k$.

\section{The \textsc{LARU} algorithm} \label{sec_LARU}

To exploit accurate predictions to reduce misses while detecting and reacting to prediction errors in real time, we present \textsc{LARU} (\underline{L}earning-\underline{A}ugmented L\underline{RU}), a deterministic learning-augmented caching algorithm that achieves both goals with an efficient design. Since the original \textsc{LRU} is strongly compatible in most cases, including prefix caching for LLM serving, \textsc{LARU} builds on \textsc{LRU} and extends it by adaptively using it as a filter for eviction candidates based on detected prediction errors.

\paragraph{Misprediction as detectable event.}

Existing prediction-error detection methods for learning-augmented caching fall into three categories. The first compares the accumulated cost of the prediction-driven policy against a robust fallback (e.g., \textsc{LRU}), as in \textsc{BlindOracle\&LRU}~\cite{wei2020better}. This requires maintaining two cache index structures to simulate both eviction processes, which can be prohibitive when the index is nontrivial (e.g., \texttt{RadixTree} in SGLang, as discussed in Section \ref{subsec:cache_index_structures}).

The second, \textsc{Guard}~\cite{chen2025robustifyinglearningaugmentedcachingefficiently}, detects mispredictions via \textit{prediction-induced misses} within \textsc{Marker}-based phases and starts a new phase only after all cached items have been requested or evicted, which makes it incompatible with prefix caching and incurs unbounded space complexity. The third recomputes the offline optimum on the observed prefix and checks whether \textsc{OPT} would also miss, as in \textsc{F\&R}~\cite{sadek2024algorithms}. While accurate, it can require $O(n\log k)$ time per eviction and $O(n)$ space after $n$ requests, making it impractical for real-time deployment.

\paragraph{Design of \textsc{LARU}.}
\textsc{LARU} employs \textit{LRU-phase-based misprediction detection}, which tracks prediction-driven evictions within each LRU phase (i.e., each $k$ distinct items requested) and bounds space overhead by $O(k)$. Lemma \ref{lemma:lru_phase_based_misprediction_detection} establishes the feasibility of this detection. This strikes a balance between efficiency and effectiveness by avoiding heavy computation and space overhead while detecting prediction errors through their direct impact on cache cost.

\begin{lemma}\label{lemma:lru_phase_based_misprediction_detection}
    If the requested item was previously evicted due to a prediction-driven eviction within the current LRU phase, then the predictions in this phase were erroneous.
\end{lemma}
\begin{proof}
    Suppose that at time $t$, the first item in this phase (denoted by $x$) satisfies this condition, meaning that $x$ was evicted by a prediction-driven eviction at time $t'$ earlier in the same phase. This implies that the number of distinct items requested in the time interval $(t', t]$ is less than $k - 1$. Hence, there exists an item $y$ that remained in the cache ans was not unrequested in $[t', t]$, so at time $t'$, the true next-request time of $y$ is later than that of $x$. Therefore, the predictions for the eviction candidates at time $t'$ are erroneous.
\end{proof}



Upon detecting a misprediction, \textsc{LARU} further applies \textit{adaptive confidence control} to dynamically adjust its reliance on predictions. This design extends heuristic filtering (HF): unlike \textsc{HALP}~\cite{song2023halp}, which fixes the candidate set size at $4$, \textsc{LARU} adjusts it on the fly, avoiding both overly conservative and overly aggressive reliance on predictions.




\begin{algorithm}[h]
\caption{\textsc{LARU}}
\label{alg:LARU}
\begin{algorithmic}[1]
    \STATE Initialize prediction table $\mathcal{P} \gets \emptyset$
    \STATE Set $\texttt{mode} \in \{\textcolor{NavyBlue}{\textsc{Sync}}, \textcolor{brown}{\textsc{Async}\}}$
     \WHILE{\textit{a request for item $x$ arrives}}
        \IF{$k$ distinct items have been requested during this phase}
            \STATE Start a new phase
            \STATE $\lambda \gets 1$
        \ENDIF 
        \IF{\textit{$x$ is not in the cache}}
            \IF{\textit{the cache is full}}
                \IF{\textit{$x$ was evicted by predictions earlier in this phase}}
                    \STATE Evict an item $z$ according to \textsc{LRU} policy
                    \STATE $\lambda \gets \lambda / 2$
                \ELSE
                    \STATE $\mathcal{L} \gets$ top-$l$ items that \textsc{LRU} would select for eviction, where $l = \max(\lfloor \lambda k \rfloor, 1)$
                    \IF{\textcolor{NavyBlue}{\textit{${mode} = \textsc{Sync}$}}}
                        \STATE \textcolor{NavyBlue}{Predict the reuse interval for each $y \in \mathcal{L}$ and update $\mathcal{P}(y)$}
                    \ENDIF
                    \STATE Evict the item $z \in \mathcal{L}$ with the largest predicted next-request time based on $\mathcal{P}$
                \ENDIF
            \ENDIF
            \STATE Load item $x$ into the cache
        \ENDIF
        \IF{\textit{\textcolor{brown}{${mode} = \textsc{Async}$}}}
            \STATE \textcolor{brown}{Asynchronously predict for item $x$ and update $\mathcal{P}(x)$}
        \ENDIF
    \ENDWHILE
\end{algorithmic}
\end{algorithm}

\paragraph{Algorithm description.} 
\textsc{LARU} (Algorithm~\ref{alg:LARU}) runs in \textit{phases}, each spanning $k$ distinct requests. On a cache miss, it checks whether the requested item was previously evicted by a prediction-driven decision in the current phase (Line 9). If so, \textsc{LARU} performs one safe \textsc{LRU} eviction and halves the confidence $\lambda$, which controls the candidate-set size for subsequent prediction-driven evictions within the phase. Otherwise, it considers only the top-$l$ items in the \textsc{LRU} queue, where $l=\max(\lfloor \lambda k \rfloor,1)$, and evicts the one with the largest predicted next-request time (Line 17). In practice, predictions can be updated either synchronously on demand (\textcolor{NavyBlue}{\textsc{Sync}}) or asynchronously (\textcolor{brown}{\textsc{Async}}), allowing predictor calls to be offloaded from the critical path.

\begin{theorem} \label{theorem:LARU_consistency_robustness}
    \textsc{LARU} is $1$-consistent and $O(k)$-robust.
\end{theorem}
\begin{proof}
    With perfectly accurate predictions, suppose upon the request of $x$, \textsc{LARU} evicts item $z$ based on predictions (Line 17) whose next request is furthest in the future in $\mathcal{L}$. Then all other $k-1$ distinct cached items will be requested before the next request to $z$, so the current phase must end and a new phase must begin before $z$ is requested again. Thus, the condition in Line~9 is never satisfied under accurate predictions. Consequently, \textsc{LARU} always evicts according to predictions, with $\lambda = 1$ and $|\mathcal{L}| = k$ holding throughout, and therefore behaves identically to the offline optimum. This completes the proof of $1$-consistency.
    
    Suppose there are $N$ phases over a sequence of $n$ requests. Since each phase contains requests to $k$ distinct items, the offline optimum \textsc{OPT} incurs at least $N-1$ cache misses in total.

    There are at most $k$ prediction-driven evictions within a phase, since when the evicted item is requested again, the \textsc{LRU}'s policy is performed (Line 10). In addition, after $k$ evictions following \textsc{LRU}'s policy, the cache must contain the $k$ most recently requested distinct items, and no further cache misses occur within the current phase. Therefore, the number of evictions within a phase is bounded by $O(k)$, and the total number of evictions is bounded by $O(Nk)$. Consequently, the algorithm is $O(k)$ competitive relative to \textsc{OPT} regardless of prediction accuracy, completing the proof of $O(k)$-robustness.
\end{proof}

Theoretically, \textsc{LARU} provides a safe fallback to near \textsc{LRU} performance under prediction errors (that is, $O(k)$ robustness), achieves optimal hit rates under perfect predictions ($1$-consistency), and has $O(n \log k)$ time complexity and incur $O(k)$ total space overhead over $n$ requests. The $\log k$ factor in the time complexity comes from selecting the item with the largest prediction (Line 17). The $O(k)$ space overhead stems from maintaining last access times, predictions, and per-phase eviction records in a hash table.


\section{Learning-Augmented Caching in Practice} \label{sec:LCR}

\begin{figure}[h]
    \centering
    \includegraphics[width=0.9\linewidth]{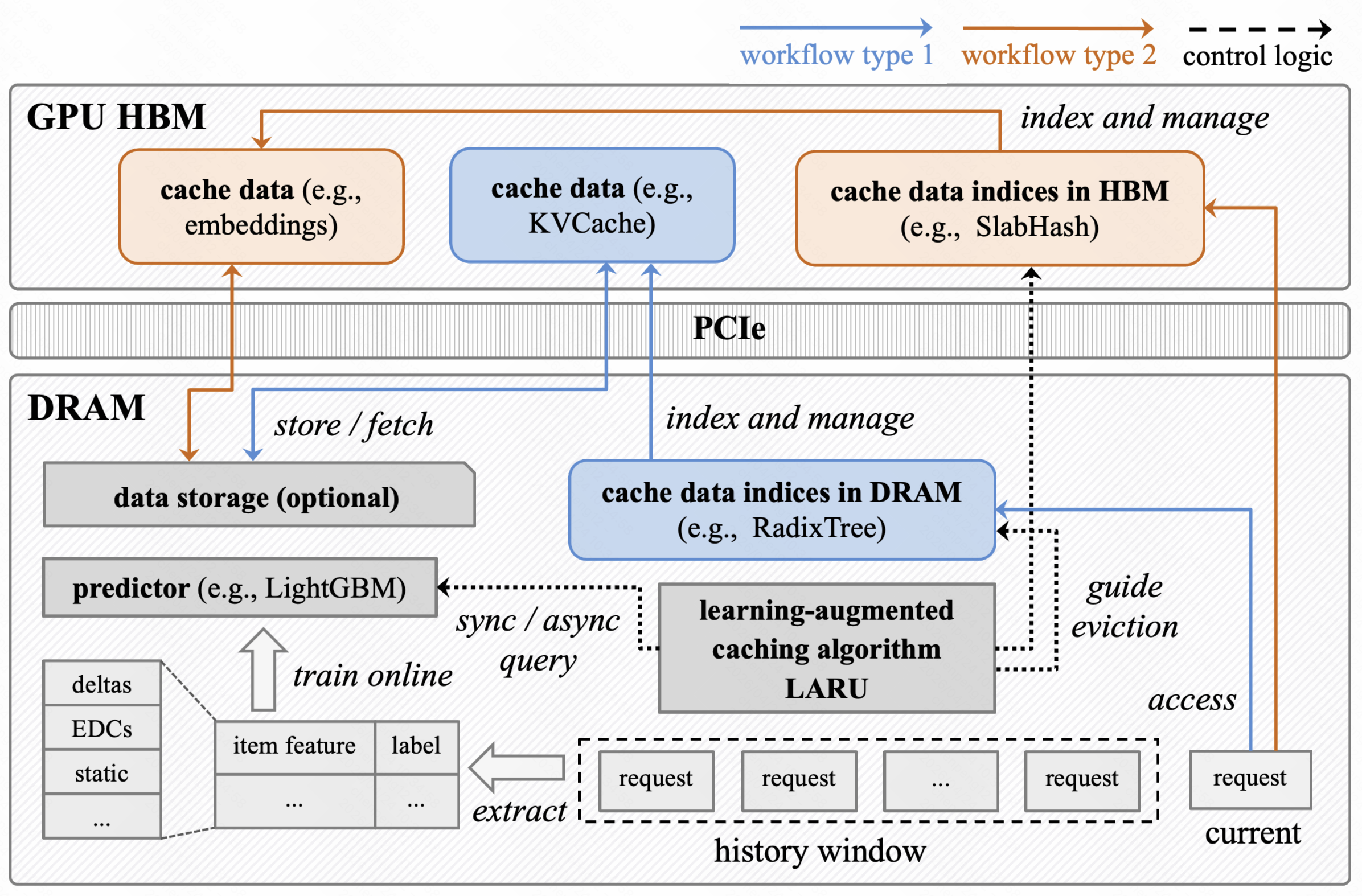}
     \caption{\textsc{LCR} architecture and two workflows based on index-structure placement.}
    \label{fig:LCR_workflow}
\end{figure}

To validate \textsc{LARU}, we present \textsc{LCR}, a learning augmented GPU cache, and integrate it into SGLang~\cite{zheng2024sglang} for LLMs and HugeCTR~\cite{wei2022gpu,10.1145/3523227.3547405} for DLRMs. \textsc{LCR} comprises three main components: a lightweight online \textit{predictor}, the \textit{\textsc{LARU} caching algorithm}, and \textit{cache index structures}. Together, they enable high hit rates under accurate predictions, robustness to noisy predictions, and high practical efficiency.

Figure~\ref{fig:LCR_workflow} presents the architecture of \textsc{LCR}. It operates as follows: incoming requests are tracked in a sliding history window, from which features and labels (e.g., next-request time) are extracted to train the predictor \textit{online}. At cache access or eviction time, \textsc{LARU} queries the predictor, either \textit{synchronously} (blocking) or \textit{asynchronously} (non-blocking), to obtain eviction priorities. These priorities are then applied by the cache indices, which manage data in HBM (e.g., KV cache and embeddings) with optional DRAM backing. The workflow depends on where the index structure is placed. Prevalent LLM inference frameworks such as SGLang and vLLM maintain indices in DRAM, whereas the well-known DLRM inference framework HugeCTR keeps indices in HBM to support efficient warp-level GPU operations under typically high request concurrency.

\subsection{LightGBM as the predictor}

A predictor for GPU caching must meet strict requirements: microsecond-level inference, minimal memory overhead, and rapid adaptation to evolving workloads, without contending for GPU compute or HBM. However, some prior learning-based caching systems use deep models (e.g., Raven~\cite{hu2022raven}, Parrot~\cite{liu2020imitation}), which add millisecond-scale latency and consume substantial HBM, competing with KV cache and embeddings. Reinforcement learning approaches (e.g., LeCaR~\cite{vietri2018driving}, \textsc{Cacheus}~\cite{rodriguez2021learning}) adapt slowly and incur high inference cost, making them ill-suited for dynamic LLM workloads.

We adopt LightGBM~\cite{ke2017lightgbm} as the predictor, a gradient-boosted decision tree that balances accuracy and efficiency. It supports fast online training (under 100 ms for tens of thousands of samples), using a sliding history window to store features from recent samples, as well as fast inference. Crucially, it runs on CPU, avoiding GPU contention. 


\subsection{Cache data index structures} \label{subsec:cache_index_structures}

\textsc{LCR}'s design is decoupled from the underlying index structure. In this paper, we focus on two representative cache index structures, \texttt{RadixTree} and \texttt{SlabHash}, used in LLM and DLRM inference, respectively. We defer the description of \texttt{SlabHash} in Appendix \ref{appendix:evaluation_dlrm}. \textsc{LCR} supports both via a common interface that abstracts insertion, lookup, and eviction primitives.

\texttt{RadixTree} is a prefix-tree structure used in SGLang~\cite{zheng2024sglang} that resides in DRAM and indexes the KV cache: prompt prefixes are represented as tree nodes, while the corresponding KV tensors are stored in GPU HBM. Each path to the root corresponds to a KV-cache prefix sequence. This enables efficient \textit{shared caching} of partial sequence prefixes, reducing the memory footprint.




\section{Evaluation for LLM Inference} \label{sec:evaluation_llm}
\subsection{Experimental Setup}
We integrate \textsc{LCR} into SGLang by replacing \texttt{RadixTree}'s default \textsc{LRU} eviction policy with \textsc{LARU}. SGLang’s \textsc{LRU} policy under prefix caching prioritizes evicting leaf nodes over internal nodes. This is consistent with the logic of \textsc{LRU}, because an internal node’s last access time is no earlier than that of any leaf in its subtree. \textsc{LARU} satisfies this requirement by always evicting an item from a candidate set consisting of leaf nodes. This is consistent with SGLang’s original policy when it falls back to \textsc{LRU}, and it does not degrade performance during prediction-driven eviction, since the next-request time of an internal node is no later than that of any leaf in its subtree.

Given the inherently low concurrency of LLM workloads (typically fewer than one hundred requests per GPU), we implement \textsc{LCR} in \textit{synchronous} mode in this case, corresponding to workflow type~1 in Figure~\ref{fig:LCR_workflow}. The time to first token (TTFT) is typically hundreds of milliseconds to several seconds, making LightGBM’s prediction latency negligible.


\paragraph{LARU and baselines.}

Following the original \texttt{RadixTree}, which aggregates many tokens into tree nodes and evicts at the node granularity, \textsc{LARU} also operates at the tree node level. We use SGLang’s original \textsc{LRU} policy as the heuristic baseline, and include \textsc{FPB} and \textsc{HF} as two additional baselines. \textsc{HF} follows \textsc{HALP}~\cite{song2023halp} and uses \textsc{LRU} to preselect four candidates for subsequent prediction-based selection.  For fairness, \textsc{FPB}, \textsc{HF}, and \textsc{LARU} all use LightGBM as the predictor. Both small and large open-source models are evaluated, including Qwen2.5-32B~\cite{team2024qwen2} and DeepSeek-R1-671B~\cite{guo2025deepseek}.



\paragraph{Predictor and hardware settings.} For each invocation, LightGBM predicts next-request times for all tree nodes in the eviction candidate set. This batch prediction is fast, under $10$~ms even with $10^4$ candidates. LightGBM uses multiple features, including the ten most recent request intervals (deltas), ten exponentially decayed counters (EDCs)~\cite{song2020learning}, the number of session turns, and the prompt length. Upon each cache request, accessing a tree node generates a training sample. We use a sliding history window of $10^5$ samples to extract features, and perform online training every $10^3$ newly collected samples. We evaluate Qwen2.5-32B on a server with four NVIDIA RTX 4090 GPUs (24 GB HBM each), and DeepSeek-R1-671B on a server with eight NVIDIA H800 GPUs (140 GB HBM each).




\paragraph{Datasets.}

We use three conversation datasets: (1) \textbf{Qwen-Bailian}: a dataset (Trace A) from publicly released traces collected in Alibaba's production environment~\cite{aliyunkvcache}, containing over 40,000 multi-turn conversational requests. (2) \textbf{Online-QA}: Company~A (anonymized for confidentiality)’s internal multi-turn conversation logs that contain 2000 conversations and 7268 requests. (2) \textbf{Aibrix-Synthetic}: a synthetic multi-turn conversational workload generated by the public Aibrix benchmark~\cite{team2025aibrix}, containing 500 conversations and 1,817 requests with a mean prompt length of 2,761 tokens.



\begin{figure}[h]
    \centering
    \includegraphics[width=1\linewidth]{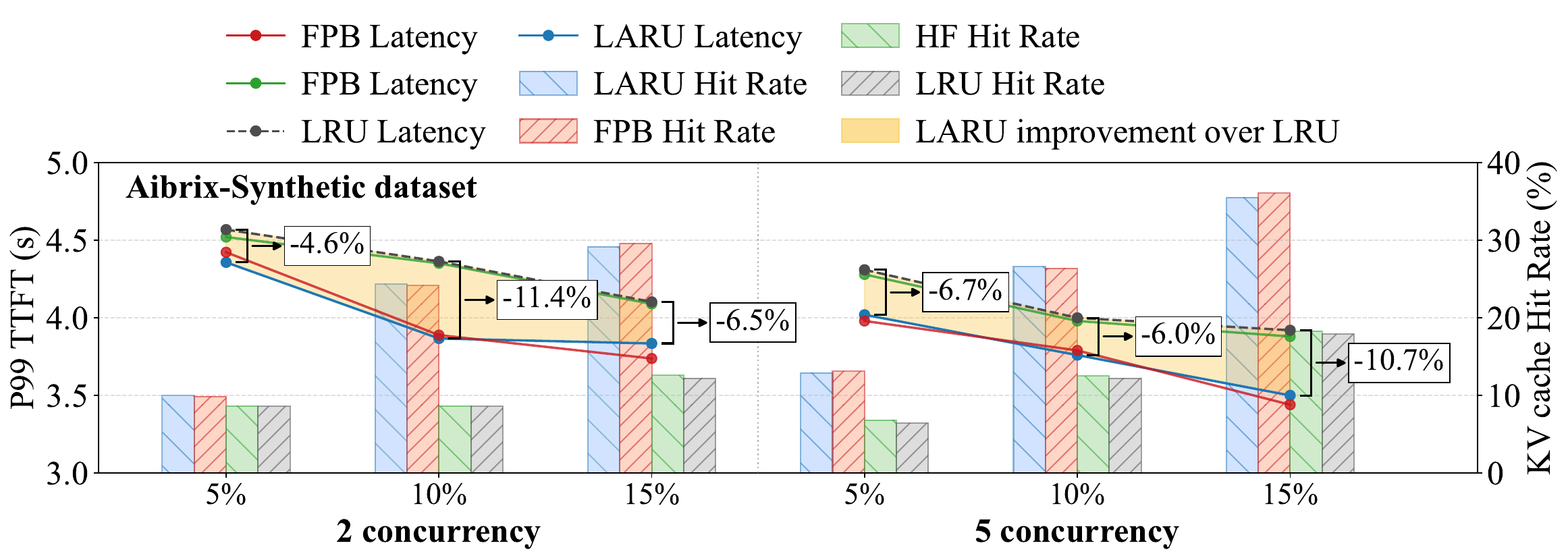}
    \caption{Performance of Qwen2.5-32B on Aibrix-Synthetic dataset under varying cache ratios and concurrency.}
    \label{fig:qwen_p99_ttft_improvement}
\end{figure}
\begin{figure}[h]
    \centering
    \includegraphics[width=1\linewidth]{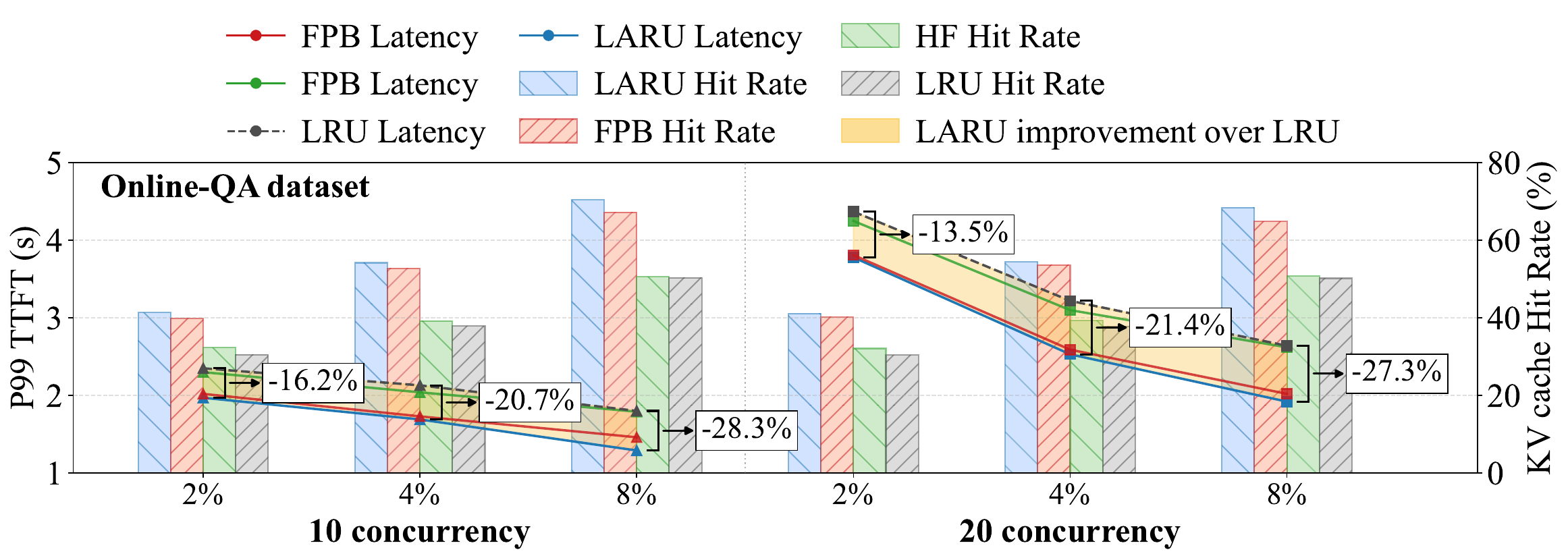}
    \caption{Performance of DeepSeek-R1-671B on Online-QA dataset under varying cache ratios and concurrency.}
    \label{fig:r1_p99_ttft_improvement}
\end{figure}

\subsection{Experimental results}
Ensuring user experience and meeting SLAs in LLM services hinges on reducing time to first token (TTFT), especially tail latency, which is largely driven by prefill efficiency during inference. Accordingly, we use \emph{P99 TTFT} as our primary metric.

On the Aibrix-Synthetic dataset with Qwen2.5-32B (Figure~\ref{fig:qwen_p99_ttft_improvement}), \textsc{LARU} reduces P99 TTFT by $6.0\%$--$10.7\%$ at cache ratios of $5\%$--$15\%$, corresponding to cache sizes of $2.7\times10^4$--$8.6\times10^4$ tokens. On the Online-QA dataset with DeepSeek-R1-671B (Figure~\ref{fig:r1_p99_ttft_improvement}), \textsc{LARU} achieves even larger gains under higher load: at cache ratios from $2\%$ to $8\%$ (cache sizes of $15.2\times10^4-46.9\times10^4$ tokens, about ), P99 TTFT improves by $13.5\%$--$28.3\%$, with consistent hit-rate increases across both $10$- and $20$-concurrency settings. Different cache ratios are used to account for differences in dataset size and device capacity constraints. These results show that \textsc{HF} yields only limited gains in this case due to its overly conservative policy, whereas \textsc{LARU} consistently improves both P99 TTFT and cache hit rate over \textsc{LRU}. In addition, \textsc{LARU} remains robust under low-accuracy predictions, as shown below.

\paragraph{Robustness verification.}
We verify the robustness of \textsc{LARU} using both real-world low-accuracy predictions and synthetic noisy predictions.

\begin{enumerate}
    \item \textbf{Real-world low-accuracy predictions}: We reduce the LightGBM predictor’s training set by setting the sliding history window to $10^2$ (0.1\% of its original size), making the predictions highly unreliable. The performance of blindly following predictions (\textsc{FPB}) deviates significantly from \textsc{LRU} in Figure~\ref{fig:r1_p99_robustness}, whereas \textsc{LARU} and \textsc{HF} remain robust and stay close to the \textsc{LRU} baseline.
    \begin{figure}[h]
    \centering
    \includegraphics[width=1\linewidth]{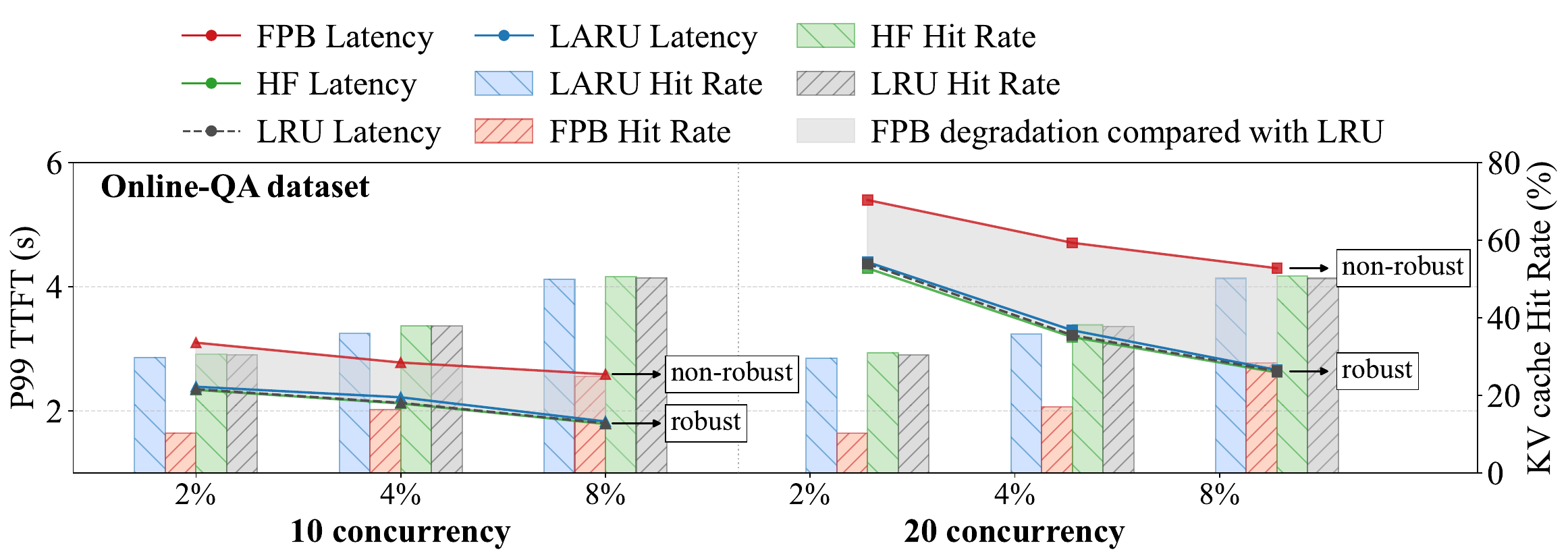}
    \caption{Performance of DeepSeek-R1-671B on the Online-QA dataset under varying cache ratios and concurrency with a low-accuracy predictor.}
    \label{fig:r1_p99_robustness}
\end{figure}
    \item 
    \begin{figure*}[h]
      \centering
      \begin{subfigure}{1\textwidth}
        \centering
        \includegraphics[width=\linewidth]{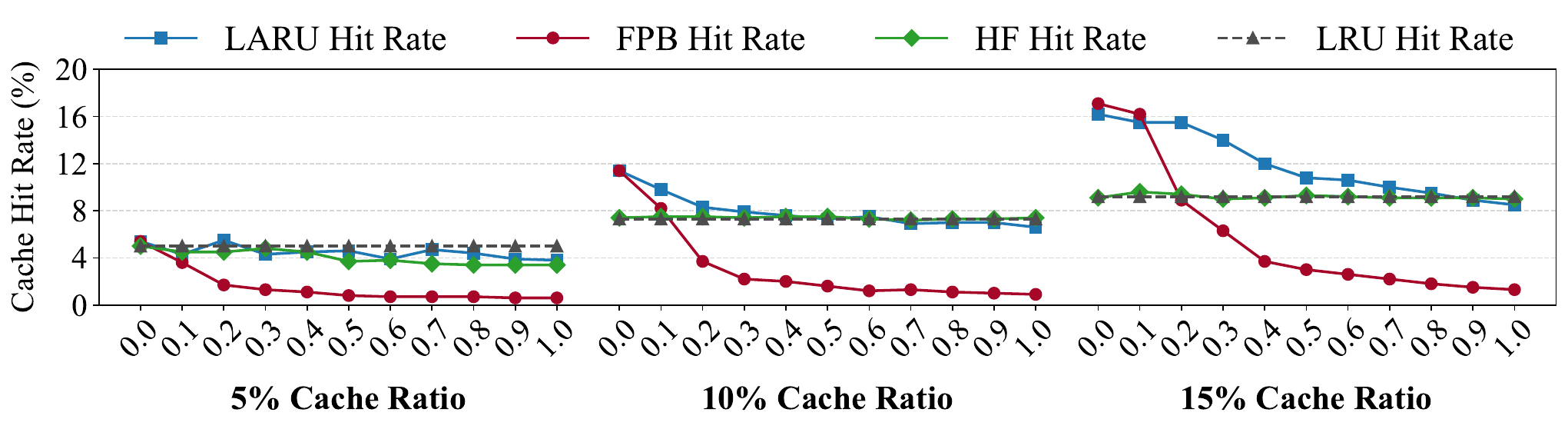}
      \end{subfigure}
    
      \vspace{-0.3em} 
    
      \begin{subfigure}{1\textwidth}
        \centering
        \includegraphics[width=\linewidth]{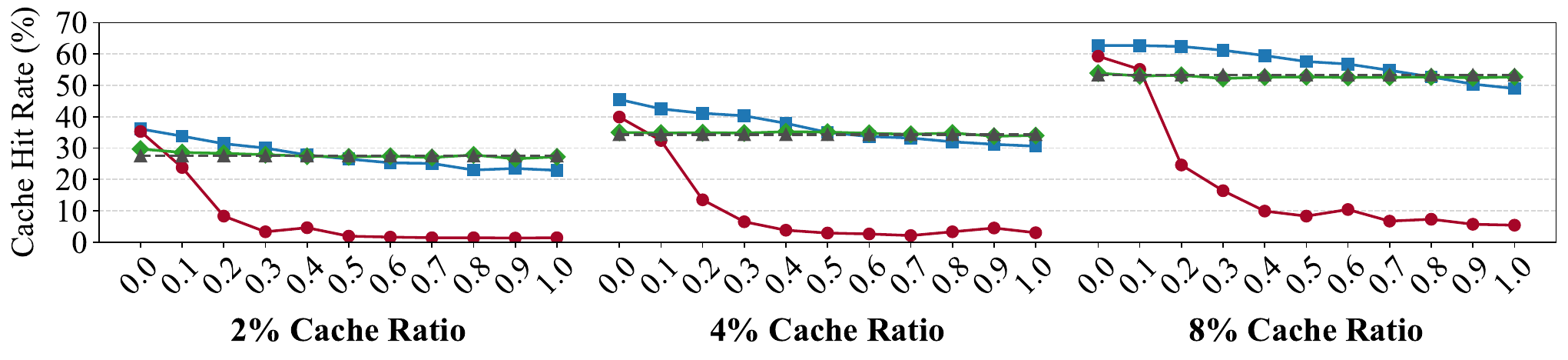}
      \end{subfigure}
    
      \vspace{-0.5em}
    
      \caption{KV cache hit rate under injected noise on Aibrix-Synthetic (top) and Online-QA (bottom). The probability of replacing predictions with worst-case noise is varied from 0.1 to 1.0.}
      \label{fig:kvcache_hit_rate_robustness}
    \end{figure*}

    \textbf{Synthetic noisy predictions}: We adopt a noise model commonly used in the learning-augmented algorithms literature for robustness evaluation \cite{lykouris2021competitive,10.1145/3528087}. Specifically, we inject controlled noise into LightGBM’s predictions to obtain prediction sets with varying accuracy: with a given probability, a prediction is replaced by the negative of the true next-request time. An error probability of $1.0$ yields completely incorrect predictions, i.e., a worst-case setting.  Figure~\ref{fig:kvcache_hit_rate_robustness} shows that \textsc{FPB} rapidly collapses as noise increases, with hit rates approaching zero even at modest error levels. \textsc{HF} remains overly conservative and fails to fully benefit from accurate predictions. In contrast, \textsc{LARU} effectively exploits accurate predictions and degrades gracefully as prediction errors increase.
    
\end{enumerate}



\section{Evaluation for DLRM Inference} \label{sec:evaluation_dlrm}

In our DLRM inference evaluation (detailed in Appendix \ref{appendix:evaluation_dlrm}), we benchmark the SparseLengthsSum (SLS) operator that is responsible for $68.9\%-73.5\%$ of DLRM runtime at realistic batch sizes using a Redis-backed DRAM embedding store and a GPU HBM cache indexed by {SlabHash} used by HugeCTR \cite{wang2022merlin}, with LightGBM predictions run asynchronously to avoid adding critical-path latency~\cite{wei2022gpu} (see workflow type~2 in Figure~\ref{fig:LCR_workflow}). Across two CTR traces (AD-CTR-User and QB-video) and varying cache ratios, \textsc{LARU} consistently lowers average SLS latency and improves hit rate: up to {19.5\%} latency reduction on AD-CTR-User and {14.2\%} on QB-video, translating to {24.2\%} and {16.6\%} higher SLS throughput, with gains increasing at larger batch sizes due to amplified DRAM random-access pressure. Under inaccurate predictions (a low-accuracy LightGBM), \textsc{FPB} degrades sharply while \textsc{LARU} remains stable and closely tracks \textsc{LRU}, confirming robustness to large prediction errors.

\section{Related Work}
Prior work spans three relevant directions: GPU caching systems for DLRMs/LLMs, learning-based caching policies, and learning-augmented algorithms with provable guarantees. Systems such as HugeCTR~\cite{wei2022gpu} and subsequent GPU caching designs~\cite{xie2022fleche, song2023ugache} demonstrate the benefits of hierarchical caching but typically rely on static heuristics (e.g., \textsc{LRU}). Learning-based policies~\cite{jain2016back, song2020learning, liu2020imitation, shi2019applying, hu2022raven, rodriguez2021learning, vietri2018driving} can improve average performance by using predictions to guide eviction, yet are often vulnerable when predictions degrade. Hybrid approaches~\cite{song2023halp, yang2023learned} restrict ML to a heuristic-filtered candidate set to improve empirical stability, while the algorithms-with-predictions literature~\cite{10.1145/3528087, wei2020better, pmlr-v80-lykouris18a, sadek2024algorithms} formalizes the trade-offs among consistency, robustness, and efficiency. We extend the discussion of related work in Appendix \ref{appendix:related_work_extended}.

\section{Conclusion} \label{sec:conclusion}

Caching in latency-critical inference frameworks, especially in industry, faces a recurring tension: fixed heuristics can perform poorly under dynamic workloads, while existing prediction-driven policies can be fragile when prediction quality degrades. Motivated by this, we investigate how to guarantee robustness using the design principles of learning-augmented algorithms, so the system remains stable and reliable even when predictions are inaccurate, a key requirement in industrial production environments.

We proposed \textsc{LARU}, a deployment-oriented learning-augmented caching algorithm, achieving practical efficiency and provable robustness when predictions are wrong. A GPU cache, \textsc{LCR}, is developed to validate \textsc{LARU} under realistic LLM and DLRM serving workloads, and extensive experiments show consistent improvements over classical baselines: on LLM workloads, it lowers P99 TTFT by as much as $28.3\%$, and on DLRMs, it reduces average SLS latency by up to $19.5\%$, while preserving stable behavior under degraded prediction quality, demonstrating that the design can both exploit accurate predictions and degrade gracefully when prediction errors increase.

Looking ahead, our work suggests a robustness-first path for bringing learning into real-world systems beyond caching: treat ML predictions as advisory signals, exposed through a small, constrained interface with bounded actions and a principled fallback, to avoid catastrophic degradation in production. This framing shifts ML for systems from best-effort optimization to risk-controlled decision making, delivering consistent gains without sacrificing stability. More broadly, it helps break the stability barrier to industrial ML adoption, making model-driven benefits more deployable and motivating future work to explore these principles across broader settings.

\bibliographystyle{unsrtnat}
\bibliography{main}

\newpage
\appendix

\section{GPU Caching in Production Systems} \label{appendix:gpu_caching_in_production_systems}

\begin{figure*}[h]
    \centering
    \includegraphics[width=1\linewidth]{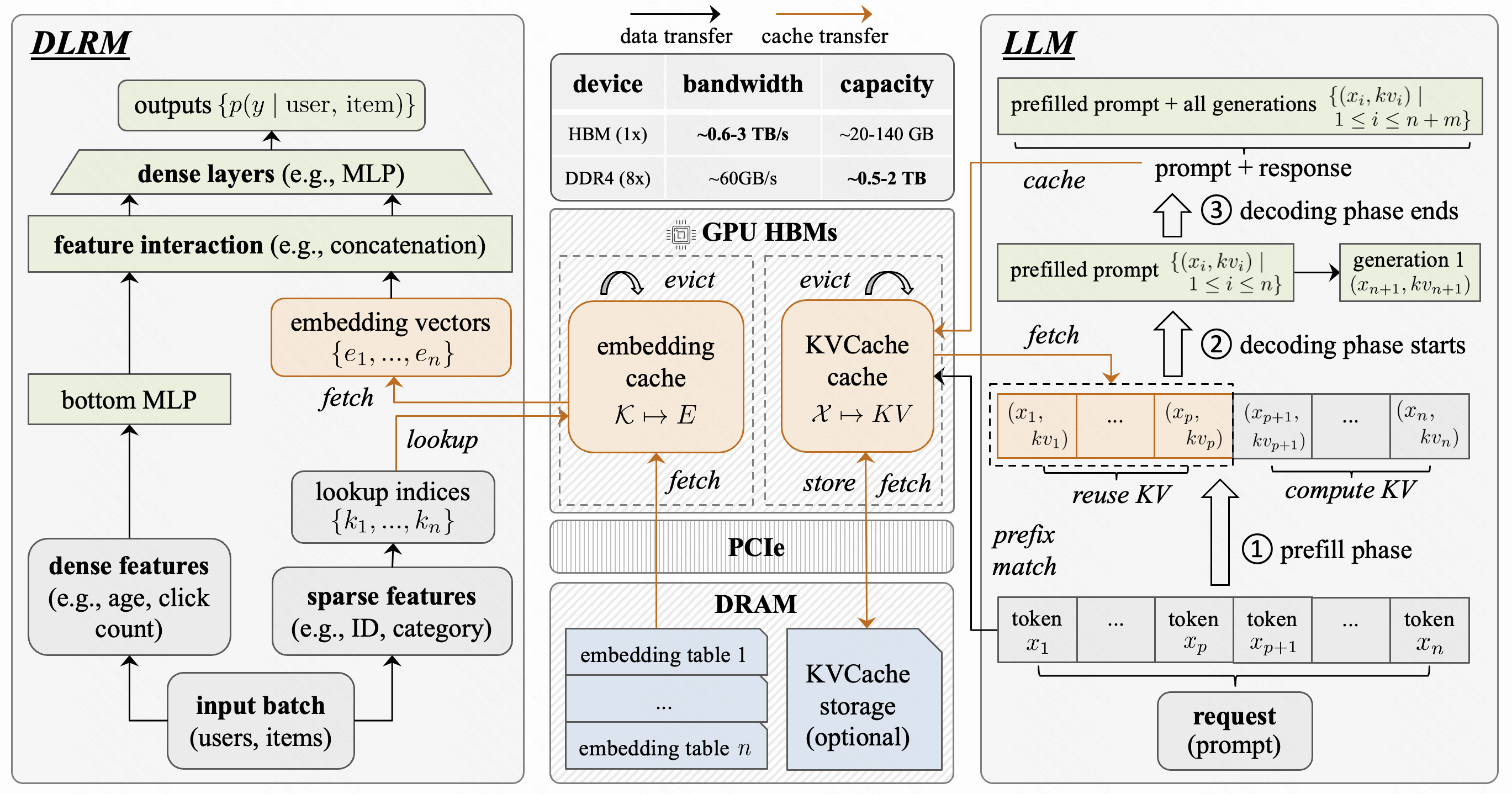}
    \caption{\textbf{Illustration of two GPU caching use cases.} (1) \textbf{DLRMs}: Inputs contain dense features and sparse features (e.g., user ID, product category). Dense features are processed by a bottom MLP, while sparse features are mapped into embedding vectors via large pre-trained tables, often stored hierarchically in GPU HBM and DRAM. (2) \textbf{LLMs}: Each request prompt is a token sequence whose KV vectors are computed during the prefill phase. If KV vectors for a prefix are cached in GPU HBM, they are reused. During decoding, the KV matrices for both the prompt and the generated tokens are cached in GPU HBM.}
    \label{fig:gpu_cache_use_cases}
\end{figure*}

In this paper, we focus on two representative model inference scenarios, LLMs and DLRMs, where effective GPU caching is critical for performance. These scenarios exhibit distinct characteristics: LLMs typically operate at lower concurrency and are highly sensitive to tail latency, especially P95 or P99 time-to-first-token (TTFT) mandated by SLAs, whereas DLRMs usually run at high concurrency where throughput is the primary metric. Despite these differences, both benefit substantially from higher hit rates in GPU-resident data. Figure~\ref{fig:gpu_cache_use_cases} illustrates the GPU cache data flow in both cases.

\paragraph{LLMs.} 
During inference, the prefill phase processes $n$ prompt tokens through self-attention, incurring $\mathcal{O}(n^2)$ compute cost per request without KV cache reuse. Without caching, each request recomputes the full attention context from scratch, making the workload strongly \textit{compute-bound}, especially for long prompts, and increasing TTFT significantly.
KV cache reuse avoids redundant computation: by storing and reusing key-value states from prior inferences, shared prefix sequences need not be recomputed. Full reuse reduces cost to $\mathcal{O}(n)$, drastically lowering TTFT. This is crucial for tail latency, as P99 delays are often driven by long-context requests such as multi-turn conversations, where prior turns are appended as prefixes to maintain \textit{contextual memory}. 
Real-world workloads like ShareGPT~\cite{sharegpt} show that 73\% of requests involve multi-turn interactions or structured prompts, exhibiting strong inter-request prefix similarity. This creates significant opportunities for cross-request KV cache sharing.
Modern frameworks such as vLLM~\cite{kwon2023efficient} and SGLang~\cite{zheng2024sglang} exploit this using \texttt{RadixTree} to manage shared prefixes efficiently, achieving up to $6.4\times$ throughput improvement. However, KV caching demands careful memory management: a single 2K-token request in an OPT-30B-sized model consumes about 2.6GB of KV cache, while model weights require over 50GB of GPU HBM. This results in tight memory budgets and places high demands on cache replacement efficiency~\cite{zheng2024sglang}. Despite these constraints, recent studies~\cite{aliyunkvcache} show that KV reuse, though skewed, exhibits strong temporal predictability.

\paragraph{DLRMs.}
Deep learning recommendation models (DLRMs) rely on massive embedding tables, often hundreds of gigabytes in size, to convert sparse categorical inputs into dense vectors. Due to limited GPU HBM capacity, only a small and frequently accessed subset can be cached on-device; the rest reside in CPU memory or remote storage. Access patterns are typically irregular and sparse, breaking memory locality and limiting data reuse. Under high concurrency, this leads to poor effective memory bandwidth from DRAM, making DLRM workloads strongly \textit{memory-bound}. Cache misses incur high latency, and system throughput is highly sensitive to embedding hit rates.
This bottleneck is well documented in both academic~\cite{kwon2019tensordimm, lee2021merci, kurniawan2023evstore, chen2024updlrm} and industrial~\cite{eisenman2019bandana, jiang2021microrec, jiang2021fleetrec, ke2020recnmp, zhou2019deep} studies. Alibaba, for instance, reported that over 60\% of inference latency in production systems stems from embedding lookups~\cite{jiang2021microrec, jiang2021fleetrec}, highlighting the importance of embedding caching. To address this, frameworks like HugeCTR~\cite{wei2022gpu,10.1145/3523227.3547405} introduce GPU-side embedding caching with \textsc{LRU}-based eviction, achieving up to $2.35\times$ throughput improvement over TensorFlow baselines at batch size 2048.
UGache~\cite{song2023ugache} improves further with a unified multi-GPU caching architecture, hotness-aware metrics, and factored extraction, achieving up to $5.25\times$ training speedup in GNNs and $3.45\times$ in recommendation inference. However, like other heuristic approaches, UGache lacks efficient predictive caching strategies.




\section{Extended Related Work} \label{appendix:related_work_extended}

\paragraph{GPU caching systems.}
Caching and memory optimization are central to improving the efficiency of large-scale recommendation and language model inference. HugeCTR~\cite{wei2022gpu} pioneered GPU-side embedding caches with hierarchical storage, achieving significant throughput gains for DLRMs. Fleche~\cite{xie2022fleche} and UGache~\cite{song2023ugache} extend this line of work by designing efficient GPU-resident or multi-GPU caching architectures. These systems demonstrate the importance of GPU caching but often rely on static heuristics such as \textsc{LRU}, which can fall short under dynamic and diverse workloads.

\paragraph{Cache algorithms with machine learning.}
A large body of work explores using machine learning to guide eviction decisions. \textsc{Hawkeye}~\cite{jain2016back} leverages predictive classification to approximate Belady’s optimal policy, while \textsc{LRB}~\cite{song2020learning} applies gradient boosting for CDN caching. \textsc{Parrot}~\cite{liu2020imitation} and \textsc{Glider}~\cite{shi2019applying} adopt deep learning models such as LSTMs and imitation learning to mimic the optimal policy. Other approaches, including Raven~\cite{hu2022raven}, \textsc{Cacheus}~\cite{rodriguez2021learning}, and LeCaR~\cite{vietri2018driving}, apply probabilistic modeling, adaptive switching, or online learning to improve eviction. Despite promising results, these methods often \emph{trust predictions blindly}, making them vulnerable to accuracy degradation.

\paragraph{Hybrid and robust designs.}
To address unreliability, hybrid methods combine heuristics with learned guidance. For example, \textsc{HALP}~\cite{song2023halp} and \textsc{MAT}~\cite{yang2023learned} preselect eviction candidates using \textsc{LRU} before applying predictors, which mitigates risk empirically. However, this conservative design may still fail to provide bounded robustness guarantees and can limit the ability to fully leverage accurate predictions. In parallel, the algorithms-with-predictions~\cite{10.1145/3528087} literature introduces formal robustness guarantees. \textsc{BlindOracle\&LRU}~\cite{wei2020better} and \textsc{PredictiveMarker}~\cite{pmlr-v80-lykouris18a} balance consistency and robustness, while \textsc{F\&R}~\cite{sadek2024algorithms} attains near-optimal guarantees at the cost of prohibitive complexity. Together, these works highlight the fundamental trade-offs among consistency, robustness, and efficiency.

\section{A Detailed Discussion of Existing Learning-Based Caching Systems} \label{appendix:existing_learning_based_caching_systems}

Table~\ref{table:ml_augmented_caching_systems} summarizes representative learning-based caching systems that use learned predictions to guide eviction decisions. A common pattern is to directly translate prediction outputs into eviction actions. For example, \textsc{Glider}~\cite{shi2019applying} trains an SVM to classify objects as cache-friendly or cache-averse and preferentially evicts cache-averse items. \textsc{LRB}~\cite{song2020learning} uses gradient boosting machines (GBM) to predict next-request time and then converts it into a binary eviction priority via thresholding. \textsc{3L-Cache}~\cite{zhou20253l} improves sampling with bidirectional traces but still largely follows GBM-predicted next-request times. Reinforcement-learning approaches~\cite{song2023halp, yang2023learned} similarly act on prediction outputs (e.g., predicting eviction actions or scores). We refer to these designs as \textsc{FPB} (\textit{Follow Prediction Blindly}): a policy that evicts the item with the highest predicted eviction priority (e.g., the farthest predicted next-request time, or the item predicted to be evicted next by an offline oracle). The core weakness of \textsc{FPB} is the lack of robustness: when predictions are inaccurate, the policy has no mechanism to limit worst-case harm, so performance may degrade sharply and can even become worse than classical heuristics.

\begin{table}[h]
\centering
\caption{\textbf{Existing representative learning-based caching systems.} The prediction column denotes the value learned by the ML model. \textit{Binary priority} indicates whether the offline optimal would evict the item next (1 if yes, 0 otherwise). \textit{Next-request time} specifies when the item will be reused. \textit{Decision} refers to the optimal eviction action.}
\label{table:ml_augmented_caching_systems}
\small
\begin{tabular}{lcccc}
\toprule
 \textbf{System} & \textbf{ML Model} & \textbf{Prediction} & \textbf{Cache Policy} \\
\midrule
\textsc{Glider} \cite{shi2019applying} & SVM & binary priority & \textsc{FPB} \\
\textsc{LRB} \cite{song2020learning} & GBM & next-request time & \textsc{FPB} \\
\textsc{3L-Cache} \cite{zhou20253l} & GBM & next-request time & \textsc{FPB} \\
\textsc{Raven} \cite{hu2022raven} & Mixture Density Network & next-request time & \textsc{FPB} \\
\textsc{Parrot} \cite{liu2020imitation} & LSTM + attention & decision & \textsc{FPB} \\
\textsc{LeCaR} \cite{vietri2018driving} & Reinforcement Learning & decision & \textsc{FPB} \\
\textsc{\textsc{Cacheus}} \cite{rodriguez2021learning} & Reinforcement Learning & decision & \textsc{FPB} \\
\textsc{HALP} \cite{song2023halp} & 1-hidden layer MLP & binary priority & \textsc{HF} \\
\textsc{MAT} \cite{yang2023learned} & GBM & next-request time & \textsc{HF} \\
\bottomrule
\end{tabular}
\end{table}

To mitigate this brittleness, more conservative systems such as \textsc{HALP}~\cite{song2023halp} and \textsc{MAT}~\cite{yang2023learned} adopt a hybrid pipeline: they first use \textsc{LRU} (or another heuristic) to pre-select a small set of eviction candidates (e.g., size $4$ \cite{song2023halp}) and then apply ML only to score and rank within this restricted set. We denote this family as \textsc{HF} (\textit{Heuristic-Filtered}). By constraining the action space exposed to ML, \textsc{HF} tends to improve empirical stability under noise, but at the expense of expressiveness: even with accurate predictions, the heuristic filter may exclude globally suboptimal candidates (or protect low-value items), thereby capping the achievable gains compared to designs that can fully exploit accurate predictions.

\section{Existing Learning-Augmented Caching Algorithms and Limitations} \label{appendix:existing_learning-augmented_caching_algorithms_and_limitations}

Recent theoretical work has advanced learning-augmented caching with robustness guarantees~\cite{mitzenmacher2022algorithms}, yet a gap remains between theory and practice. These algorithms are evaluated on \textit{consistency} (performance under perfect predictions) and \textit{robustness} (worst-case performance under large errors). But real-world systems demand a third dimension: \textit{efficiency}, which captures time and space complexity as well as implementation overhead. Table~\ref{table:alps_comparisons} lists representative caching algorithms as well as \textsc{LARU}.

\begin{table}[h]
\centering
\caption{Comparison of performance guarantees among representative learning-based caching algorithms. Let $n$ denote the total number of requests, and $k$ the maximum number of stored items.}
\label{table:alps_comparisons}
\small
\begin{tabular}{lccccc}
\toprule
 \textbf{Randomized Algorithm} & \textbf{Consistency} & \textbf{Robustness} & \textbf{Time Complexity} & \textbf{Space Complexity} \\
\midrule
\textsc{FPB} & $1$ & $\infty$ & $O(\log k)$ & $O(k)$ \\
\textsc{PredictiveMarker} \cite{pmlr-v80-lykouris18a} & $2$ & $O(\log k)$ & $O(\log k)$ & $O(k)$ \\
\textsc{LMarker} \cite{rohatgi2020near} & $4$ & $O(\log k)$ & $O(\log k)$ & $O(k)$ \\
\textsc{LNonMarker} \cite{rohatgi2020near} & $4$ & $\infty$ & $O(\log k)$ & $O(k)$ \\
\textsc{F\&R} \cite{sadek2024algorithms} & $1$ & $O(\log k)$ & $O(n \log k)$ & $O(n + k)$ \\
\textsc{Guard} \cite{chen2025robustifyinglearningaugmentedcachingefficiently} & $1$ & $O(\log k)$ & $O(\log k)$ & $\infty$ \\
\midrule
\textbf{Deterministic Algorithm} & \textbf{Consistency} & \textbf{Robustness} & \textbf{Time Complexity} & \textbf{Space Complexity} \\
\midrule
\textsc{BlindOracle\&LRU} \cite{wei2020better} & $2$ & $O(k)$ & $O(\log k)$ & $O(k)$ \\
\textsc{LARU} (this work) & $1$ & $O(k)$ & $O(\log k)$ & $O(k)$ \\
\bottomrule
\end{tabular}
\end{table}

\textsc{FPB} achieves ideal $1$-consistency by following prediction blindly, matching the offline optimum when predictions are accurate. However, it has no bounded robustness. \textsc{BlindOracle\&LRU} \cite{wei2020better} improves robustness to $2k$ by switching to \textsc{LRU} under poor predictions, but its consistency is 2, limiting hit rate even under perfect predictions. Note that $k$ is the maximum item number in the cache. In addition, it requires maintaining multiple parallel caches. This is prohibitive in GPU systems where metadata access and synchronization are expensive, especially with hierarchical structures like \texttt{RadixTree} in SGLang~\cite{zheng2024sglang}. Other algorithms face different trade-offs. \textsc{PredictiveMarker}~\cite{pmlr-v80-lykouris18a} achieves $4H_k$-robustness by augmenting 
\textsc{Marker}~\cite{fiat1991competitive}, but its consistency is also capped at $2$. Here $H_k = \sum_{i=1}^k \frac{1}{i} \in [\ln (k+1) , \ln k + 1]$. \textsc{LNonMarker}~\cite{rohatgi2020near} lacks robustness bounds. \textsc{F\&R}~\cite{sadek2024algorithms} achieves ideal $1$-consistency and $O(\log k)$-robustness, but its amortized $O(n \log k)$ per-request complexity is infeasible for real-time systems. Here $n$ is the total number of page requests. In contrast, practical algorithms achieve $O(\log k)$ per-access cost by maintaining a priority queue over cached items.



\section{Evaluation for DLRM Inference} \label{appendix:evaluation_dlrm} 

\subsection{Experimental Setup}

To isolate caching effects, we build an SLS operation benchmark integrated with \textsc{LCR} that decouples the SparseLengthsSum (SLS) operator from HugeCTR~\cite{wei2022gpu}, using Redis as a DRAM-based backend. SLS is a core primitive in DLRMs, and production-scale studies~\citep{ke2020recnmp} show it dominates inference latency at realistic batch sizes, accounting for 68.9\%-73.5\% of total runtime. It follows a gather-reduce pattern, $\textit{Output}=\textit{SLS}(\textit{Emb}, \textit{Indices}, \textit{Lengths})$, where $\text{Emb}$ is an $[\text{Emb size}, \text{vector size}]$ table, $\text{Indices}$ specifies which embeddings to fetch, and $\text{Lengths}$ gives the number of indices per sample. The fetched vectors are reduced into shape $[\text{batch size}, \text{vector size}]$.


We use 128-float embedding vectors and set the pooling size to $50$, following common industry practice and prior work~\citep{ke2020recnmp}. Full tables reside in DRAM, while a subset is cached in GPU HBM.

\paragraph{Cache index structure.} The cache index uses HugeCTR’s \texttt{SlabHash}~\cite{ashkiani2018dynamic} in HBM, a GPU-optimized hash table for indexing embedding vectors in DLRM inference. It organizes memory into fixed-size slots grouped into \textit{slabs} and \textit{slabsets}, enabling \textit{warp-synchronous probing} and batched updates. This design improves memory coalescing and reduces bank conflicts, thereby achieving high throughput. Items are first hashed to a slabset; each slabset contains two slabs, each with 32 slots for warp-level efficiency. The caching algorithm then operates locally within each slabset.




\paragraph{LARU and baselines.} \textsc{LARU} uses hash tables to record prediction-driven evictions in slabsets, with total space overhead of $O(k)$. The total size of these hash tables is less than 0.2\% of the cache size, which is negligible. We use HugeCTR’s policy, \textsc{LRU}, as the heuristic baseline. \textsc{FPB} (\textit{Follow Prediction Blindly}) serves as the learning-augmented baseline, commonly used in prior learning-based caching systems such as \textsc{LRB}~\cite{song2020learning}, \textsc{3L-Cache}~\cite{zhou20253l}, and \textsc{Raven}~\cite{hu2022raven}. \textsc{HF} is another baseline that implements a heuristic-filtered strategy, adopted by \textsc{HALP}~\cite{song2023halp} and \textsc{MAT}~\cite{yang2023learned}.


%

\paragraph{Predictor and Hardware settings.} 

Due to high DLRM concurrency, learning-augmented algorithms run LightGBM in \textit{asynchronous mode} to keep prediction off the critical path. LightGBM uses four lightweight features (timestamp, key, request count, and last inter-request interval) and triggers a background prediction. This corresponds to workflow type 2 in Figure~\ref{fig:ANCHOR_workflow}. LightGBM predicts quickly; for example, at batch size $512$, it takes under $1$ms, far below the SLS latency. Experiments run on a bare-metal server with two NVIDIA A10 GPUs and 512GB DDR4 DRAM. The GPUs provide $\sim$600,GB/s memory bandwidth, while DRAM offers $\sim$60,GB/s random-read bandwidth.



\paragraph{Datasets.}
We evaluate on two real-world CTR datasets: an internal trace from a leading global short-video platform (anonymized as Company~A) and a public dataset from the Tenrec benchmark~\cite{yuan2022tenrec}, released by a major global social networking company. (1) \textbf{AD-CTR-User} contains user-ID embedding queries from Company~A’s production DLRMs for advertisement recommendation, representing large-scale industrial deployment. (2) \textbf{QB-video} is a Tenrec \cite{yuan2022tenrec} subset from Tencent’s video recommendation platform, containing millions of user–item interactions with positives, implicit negatives, and auxiliary signals (e.g., category and time).


\subsection{Experimental results}

\begin{figure*}[h]
  \centering
  \begin{subfigure}{1.0\textwidth}
    \centering
    \includegraphics[width=\linewidth]{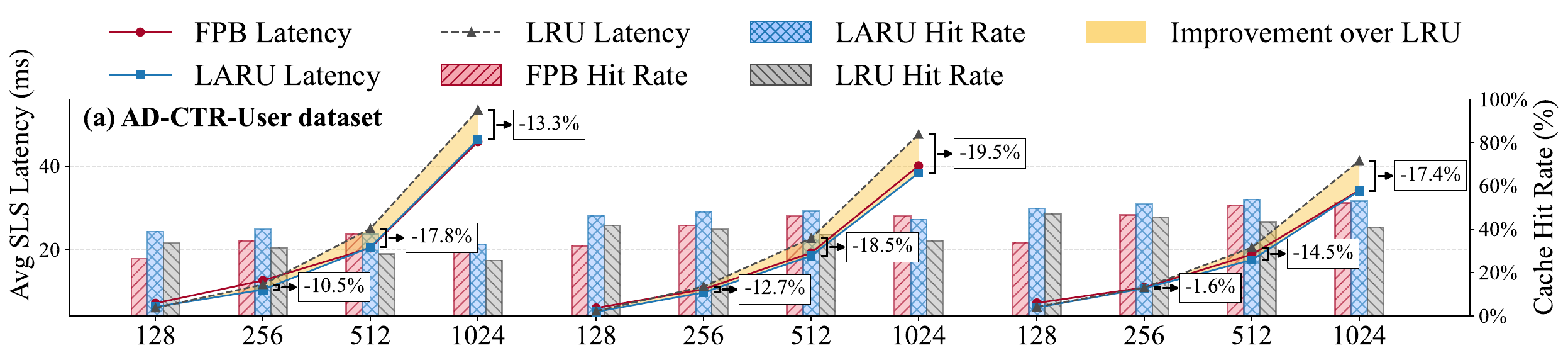}
  \end{subfigure}

  \vspace{-0.5em} 

  \begin{subfigure}{1.0\textwidth}
    \centering
    \includegraphics[width=\linewidth]{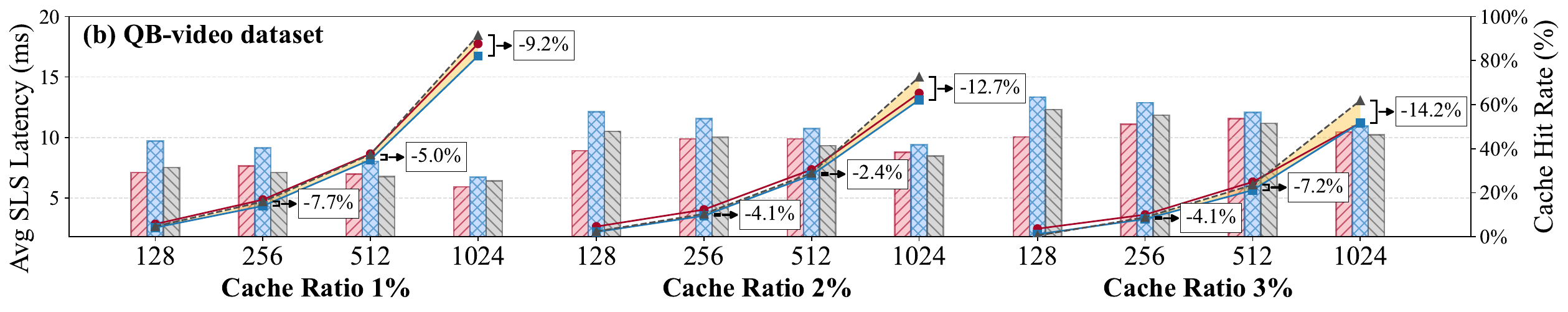}
  \end{subfigure}

  \vspace{-0.5em} 

  \caption{\textbf{Average SLS latency improvement across different cache ratios and batch sizes.} Results are reported for (a) AD-CTR-User and (b) QB-video datasets, evaluated at cache ratios of $1\%$, $2\%$, and $3\%$. Numbers in boxes denote \textsc{LARU}’s improvement over \textsc{LRU}. The cache ratio is the cache size over the total number of distinct items.}
  \label{fig:sls_latency_improvement}
\end{figure*}

We evaluate SLS performance using \emph{average latency}, consistent with HugeCTR~\cite{wei2022gpu}, which is key to improving SLS throughput. Since real-world embedding tables often exceed several TBs, we limit cache ratios to at most $3\%$. Figure~\ref{fig:sls_latency_improvement} shows that, across both datasets, \textsc{LARU} consistently lowers latency and improves hit rates. On AD-CTR-User, it reduces average latency by up to 19.5\%, with the largest gains at high batch sizes. On QB-video, reductions reach 14.2\%, translating to 24.2\% and 16.6\% higher SLS throughput, respectively. These gains grow with batch size as concurrency increases DRAM random accesses. Overall, \textsc{LARU} substantially outperforms \textsc{LRU} at the large batch sizes common in production environments and performs close to \textsc{FPB}, indicating it can effectively leverage accurate predictions.

\paragraph{Robustness Verification.} 

\begin{figure*}[h]
  \centering
  \begin{subfigure}{1\textwidth}
    \centering
    \includegraphics[width=\linewidth]{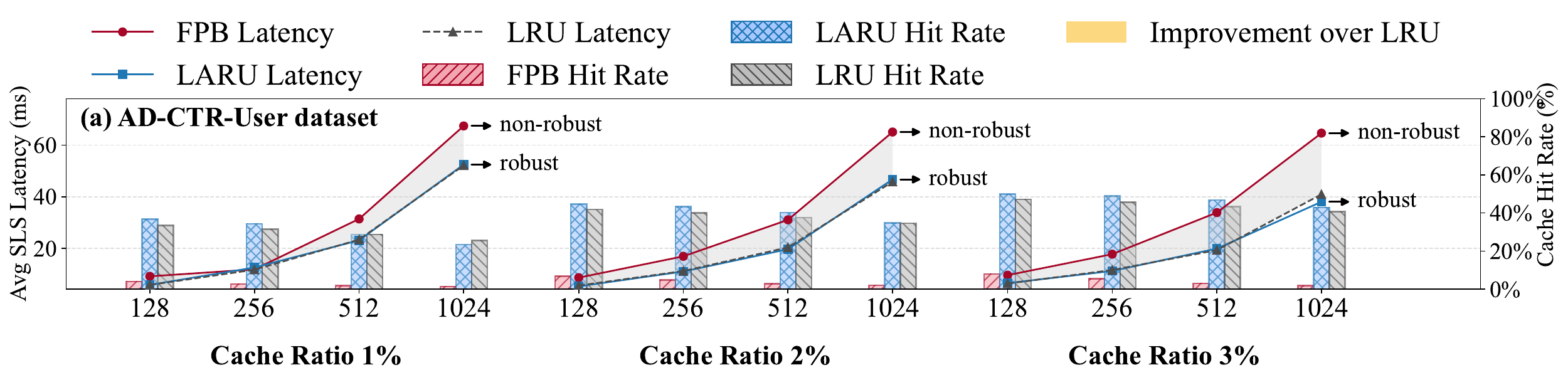}
  \end{subfigure}

  \vspace{-1.5em} 

  \begin{subfigure}{1\textwidth}
    \centering
    \includegraphics[width=\linewidth]{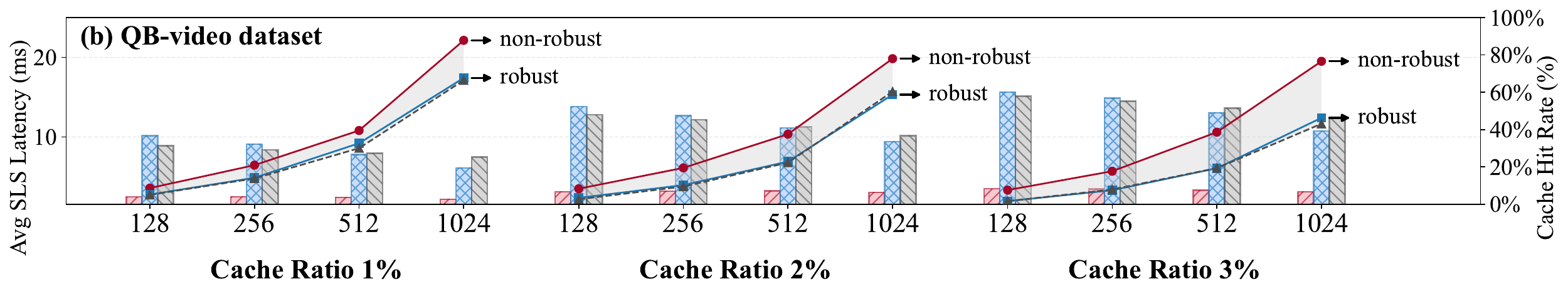}
  \end{subfigure}

  \vspace{-0.3em}

  \caption{\textbf{Robustness Verification.} Average SLS latency under different batch sizes and cache ratios on (a) AD-CTR-User and (b) QB-video datasets. Both \textsc{FPB} and \textsc{LARU} use the same low-accuracy LightGBM predictor to assess robustness.}
  \label{fig:sls_robustness}
\end{figure*}

When predictions are inaccurate, \textsc{LARU}’s robustness takes effect. We test this with a weak LightGBM predictor trained on a single feature and limited to one tree at inference time, yielding very low accuracy. Figure~\ref{fig:sls_robustness} shows that \textsc{FPB} degrades sharply under these predictions, especially at larger batch sizes, while \textsc{LARU} remains stable across cache ratios, batch sizes, and workloads, closely tracking \textsc{LRU}. This confirms that \textsc{LARU} preserves performance and stability in \textsc{LCR} even with large prediction errors.

\section{Experimental Implementation Details}
The supplementary materials include all code and non-confidential traces needed to reproduce the experimental results presented in the paper. We integrate \textsc{LCR} into two open-source inference frameworks, one for LLM inference and the other for DLRM inference.


\begin{enumerate}
    \item \textbf{\textsc{LCR-on-SGLang}.} For LLMs, we integrate \textsc{LCR} into SGLang (version 0.4.9.post2) by adding LightGBM support, implementing an online training framework, and replacing the default \texttt{RadixTree} cache implementation with a new cache that supports \textsc{LARU}. This integration enables systematic comparison of different KV cache policies.
    
    \item \textbf{\textsc{SLS-Cache-Bench}.} A dedicated benchmark that isolates the SparseLengthsSum (SLS) operation in DLRM inference from the HugeCTR framework and integrates \textsc{LCR}. The benchmark supports direct comparison of GPU embedding cache policies. We implement an alternative GPU cache in the \texttt{flash\_cache} directory while retaining the original cache indices and \texttt{SlabHash}, and enable host-to-device transfer of predictions to realize the logic of \textsc{LARU}. The benchmark is self-contained, reproducible, and readily extensible for future evaluations.
    
\end{enumerate}

The hardware used to generate the results in this paper is described in Section \ref{sec:evaluation_llm} for LLM inference and Section \ref{appendix:evaluation_dlrm} for DLRMs inference.

\end{document}